\documentclass[letterpaper]{article} % DO NOT CHANGE THIS
\usepackage[submission]{aaai25}  % DO NOT CHANGE THIS
\usepackage{times}  % DO NOT CHANGE THIS
\usepackage{helvet}  % DO NOT CHANGE THIS
\usepackage{courier}  % DO NOT CHANGE THIS
\usepackage[hyphens]{url}  % DO NOT CHANGE THIS
\usepackage{graphicx} % DO NOT CHANGE THIS
\usepackage{natbib}  % DO NOT CHANGE THIS AND DO NOT ADD ANY OPTIONS TO IT
\usepackage{caption} % DO NOT CHANGE THIS AND DO NOT ADD ANY OPTIONS TO IT
\usepackage{algorithm}
\usepackage{algorithmic}

\usepackage{amsmath}
\usepackage{amssymb}
\usepackage{mathtools}
\usepackage{amsthm}
\usepackage{microtype}
\usepackage{subfigure}
\usepackage{booktabs} % for professional tables
\usepackage{multirow}
\usepackage{pythonhighlight}
\usepackage{paralist}
\usepackage{arydshln}
\usepackage{pifont}

\usepackage{newfloat}
\usepackage{listings}
\DeclareCaptionStyle{ruled}{labelfont=normalfont,labelsep=colon,strut=off} % DO NOT CHANGE THIS
\floatstyle{ruled}
\newfloat{listing}{tb}{lst}{}
\floatname{listing}{Listing}
\title{ExtremeCast: Boosting Extreme Value Prediction for Global Weather Forecast}
\author{
    Wanghan Xu\textsuperscript{\rm 1 \rm 2}
    Kang Chen\textsuperscript{\rm 1 \rm 3}
    Tao Han\textsuperscript{\rm 1 \rm 4}
    Hao Chen\textsuperscript{\rm 1}
    Wanli Ouyang\textsuperscript{\rm 1}
    Lei Bai\textsuperscript{\rm 1}
}
\affiliations{
    \textsuperscript{\rm 1}Shanghai Artificial Intelligence Laboratory\\
    \textsuperscript{\rm 2}Shanghai Jiao Tong University\\
    \textsuperscript{\rm 3}University of Science and Technology of China\\
    \textsuperscript{\rm 4}The Hong Kong University of Science and Technology\\
}

\usepackage{bibentry}
\begin{document}

\maketitle

\begin{abstract}
Data-driven weather forecast based on machine learning (ML) has experienced rapid development and demonstrated superior performance in the global medium-range forecast compared to traditional physics-based dynamical models. However, most of these ML models struggle with accurately predicting extreme weather, which is related to training loss and the uncertainty of weather systems. Through mathematical analysis, we prove that the use of symmetric losses, such as the Mean Squared Error (MSE), leads to biased predictions and underestimation of extreme values. To address this issue, we introduce \textbf{Exloss}, a novel loss function that performs \textbf{asymmetric optimization} and highlights extreme values to obtain accurate extreme weather forecast. Beyond the evolution in training loss, we introduce a \textbf{training-free} extreme value enhancement module named \textbf{ExBooster}, which captures the uncertainty in prediction outcomes by employing multiple random samples, thereby increasing the hit rate of low-probability extreme events. Combined with an advanced global weather forecast model, extensive experiments show that our solution can achieve state-of-the-art performance in extreme weather prediction, while maintaining the overall forecast accuracy comparable to the top medium-range forecast models. Code and the model checkpoints are available at \textcolor{blue}{\textbf{https://github.com/black-yt/ExtremeCast}} .
\end{abstract}


\section{Introduction}
Weather forecast plays a crucial role in society and impacts various industries including energy ~\cite{meenal2022weather}, agriculture ~\cite{fathi2022big}, and transportation ~\cite{wang2020evaluation}. In recent years, the field of Numerical Weather Prediction has witnessed the rapid development of many data-driven forecast models based on machine learning (ML) ~\cite{kurth2023fourcastnet, bi2023accurate, chen2023fengwu}, which are usually trained on extensive meteorological data spanning decades, achieving remarkable accuracy and fast inference.

\begin{figure}[t]
\centerline
{\includegraphics[width=8.6cm]{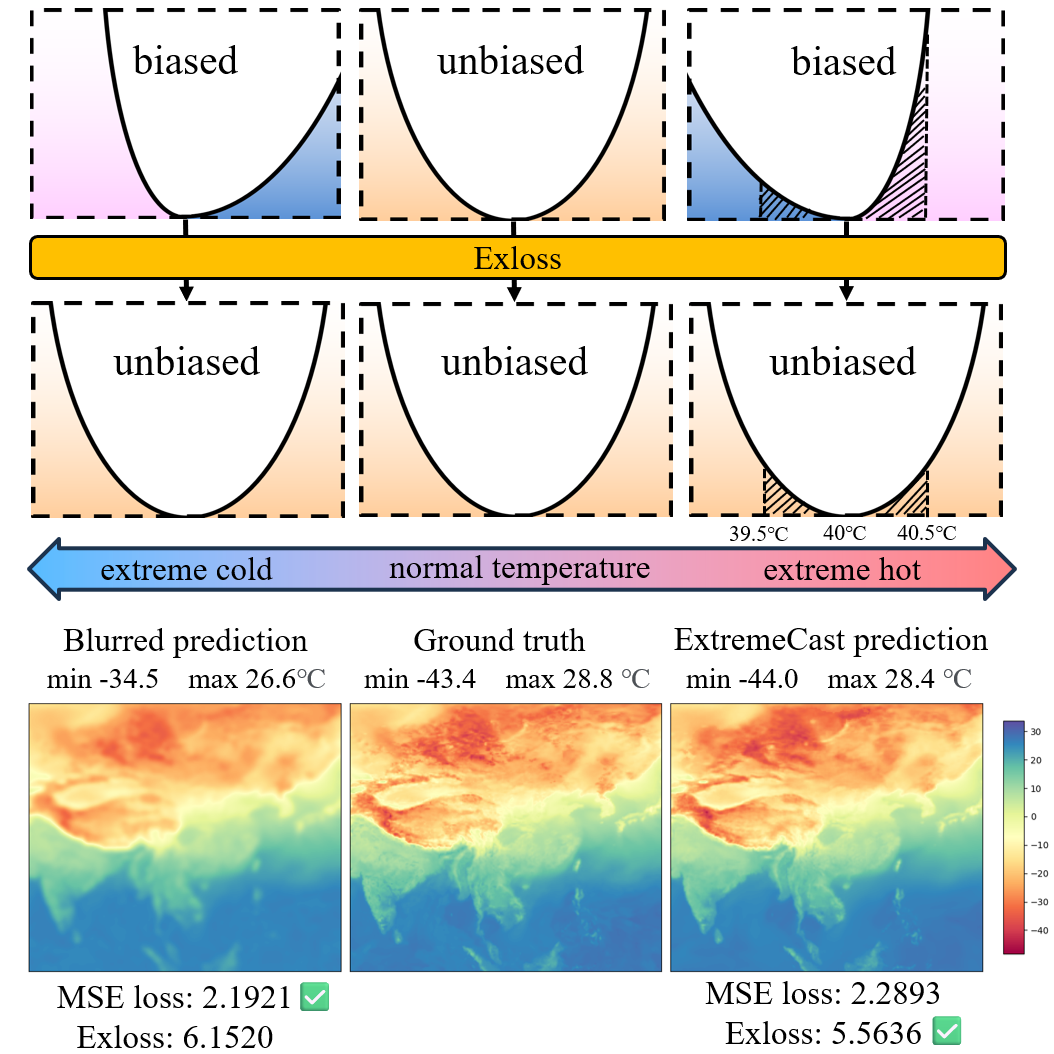}}
\vspace{-0.5em}
\caption{\textbf{MSE loss vs. Exloss.} Through theoretical analysis, we found that MSE loss will give underestimated extreme value predictions. An asymmetric loss function, Exloss, is designed to address this bias by balancing the data distribution.}
\label{fig: Exloss}
\vspace{-1.6em}
\end{figure}

Despite the advancements in ML forecast models, there remains a significant challenge when it comes to the extreme weather forecast. More specifically, these models tend to underestimate extreme values and produce smooth outputs~\cite{gong2024cascast}, especially when the prediction lead time extends, which greatly hinders the prediction of natural disasters like typhoons and heatwaves, as shown in Figure \ref{fig: Exloss}. Some previous works~\cite{lopez2023global, ni2023kunyu} have highlighted the notable influence of training loss on extreme value prediction, and empirically improve extreme values by adjusting the training loss. However, these work generally lack theoretical analysis and targeted methodologies. In addition, modifying the loss alone cannot capture the inherent uncertainty in the atmosphere, i.e., identical initial conditions may lead to multiple potential outcomes, and certain extreme events with low probabilities only appear in specific outcomes. The methodology for capturing extreme values in uncertain systems remains unexplored.

In this paper, we present a theoretical explanation from extreme value theory (EVT)~\cite{smith1990extreme} to illustrate why models utilizing Mean Squared Error (MSE) loss perform poorly in extreme weather prediction. Through mathematical analysis, we found that MSE loss is effective only for normally distributed data, in the case of asymmetric extreme value distributions, it emphasizes minimizing the overall average error rather than forecasting extreme values accurately, leading to an underestimation bias. Building on this analysis, we propose an \textbf{asymmetric} loss function called \textbf{Exloss}, which incorporates a quantitative scaling function that adjusts the distribution of asymmetric extreme values towards a more balanced distribution, correcting the model's underestimate tendency, as illustrated in Figure \ref{fig: Exloss}.

Moreover, due to the inherent uncertainty of the atmospheric system, a single prediction outcome may not encompass all extreme events. Inspired by ensemble algorithms~\cite{gneiting2005weather}, we introduce \textbf{ExBooster}, which generates multiple potential prediction outcomes via random sampling and subsequently employs the rank histograms algorithm ~\cite{hamill2001interpretation} to amalgamate all prediction results. This module serves as a plug-and-play component that does not necessitate training but significantly enhances the incorporation of extreme values in predictions.

Building upon an advanced medium-range global weather forecast model, i.e., FengWu~\cite{chen2023fengwu}, and a cascaded diffusion model~\cite{ho2020denoising}, we come up with ExtremeCast, a robust and powerful global weather forecast model that works on the $0.25^{\circ }$ high-resolution. Trained with large-scale global atmosphere reanalysis dataset, i.e., ERA5~\cite{hersbach2020era5}, ExtremeCast achieves state-of-the-art (SOTA) performance on extreme value metrics compared with both the top-performing data-driven global weather forecast models and the physics dynamic model, while maintaining the competitive overall accuracy. We summarize the contributions of this paper as follows:

\begin{itemize}
\item From the perspective of extreme value theory, we propose a theoretical explanation for why MSE-based model is difficult to predict extreme values and prove that MSE leads to underestimation of extreme values. 
\item We design a novel loss function named \textbf{Exloss} based on an asymmetric design, which balances data distribution to eliminate the bias of the above underestimated forecasts and provide accurate extreme weather forecasts.
\item We introduce \textbf{ExBooster}, a training-free module that models atmospheric uncertainty through random samplings, effectively boosting the forecast hit rate of extreme values.
\item Large-scale experiments for global weather forecast on the $0.25^{\circ }$ resolution show that our method significantly improve the forecasting ability of extreme weather, which is supported by obtaining better performance on extreme value metrics (e.g., RQE and SEDI) while maintaining the overall accuracy metrics (e.g., RMSE).
\end{itemize}

\section{Related Work}

\subsection{Medium-range Weather Forecast}
Medium-range weather forecast provides forecasts of weather conditions including temperature, wind speed, humidity, and more, for the next few days to two weeks. Traditional medium-range forecast models are mainly physics-based dynamic models such as the Integrated Forecasting System (IFS) from European Centre for Medium-Range Weather Forecasts (ECMWF) ~\cite{model2003ifs}, whose development is primarily limited by the computational cost of numerical models ~\cite{ben2023rise}.

In recent years, data-driven models based on machine learning show significant potential for weather forecast ~\cite{de2023machine}. FourCastNet ~\cite{kurth2023fourcastnet} utilizes Adaptive Fourier Neural Operator networks ~\cite{guibas2021adaptive} for prediction and performs a two-step finetuning to improve the accuracy of autoregressive multi-step forecast, where the model's output is fed back as input for predicting the next step. This was followed by the Pangu-Weather ~\cite{bi2023accurate}, which uses 3D Swin-Transformer ~\cite{liu2021swin} and proposes hierarchical temporal aggregation aimed at reducing iterations in autoregressive forecast through the integration of multiple models. Subsequently, GraphCast ~\cite{lam2023learning} uses graph neural networks and performs a 12-step autoregressive finetuning. FengWu ~\cite{chen2023fengwu} treats the prediction problem as a multi-task optimization and introduces a novel finetune strategy named replay buffer. FuXi ~\cite{chen2023fuxi} reduces the cumulative error of the multi-step forecast by cascading three U-Transformer models ~\cite{petit2021u} optimized at different lead times.

Although many ML models outperform ECMWF-IFS on the root mean square error (RMSE) metric, most of them suffer from the prediction smoothing issue and lag behind ECMWF-IFS on the extreme value metrics.

\subsection{Extreme Weather Forecast}

The application of ML models in predicting extreme weather remains relatively unexplored. Most existing models for extreme weather prediction are limited to regional scale or low resolution ($> 0.25^{\circ }$) rather than global scale and high resolution ($0.25^{\circ }$). Zhao et al. ~\cite{zhao2003detecting} use wavelet transform ~\cite{farge1992wavelet} to predict the occurrence of extreme weather, but cannot predict their actual values. Porto et al. ~\cite{porto2022machine} combine multiple models to address the challenge of learning diverse extreme weather patterns, which introduces several times the additional training cost. Annau et al. ~\cite{annau2023algorithmic} propose to separate high and low-frequency in the data through convolution and enhance the learning of high-frequency to improve extreme value prediction. However, the learning of high-frequency is notably more challenging than low-frequency, which has not been effectively solved. Morozov et al. ~\cite{morozov2023cmip} apply bias correction on extreme values through quantile regression, whose main limitation is that it only performs regression on $7$ quantiles, which may not sufficiently capture the continuous distribution of the data.

Some work improves extreme values by adjusting loss. Lopez-Gomez et al. ~\cite{lopez2023global} increase the learning weight of extreme values by designing an exponential-based loss function. However, this method may cause exponential explosion in practical applications \cite{wang2019better}. Ni et al. ~\cite{ni2023kunyu} identify that using GAN loss ~\cite{creswell2018generative} can enhance the performance of the model in predicting extreme values, which is relatively difficult to optimize ~\cite{berard2019closer}. 

For the global scale and high resolution extreme weather forecast, FuXi-Extreme ~\cite{zhong2023fuxi} uses a diffusion model to improve prediction accuracy of extreme weather. Building on this previous work, we use a cascaded diffusion model as a baseline component of our model and design Exloss and ExBooster to achieve SOTA performance.

\section{Method}
\subsection{Preliminary}
Formally, we denote the weather state at time period $i$ as a tensor $\mathcal X^i \in \mathbb{R} ^{C\times H\times W}$ , where $C$ represents the number of atmospheric variables, $H$ and $W$ are the height and width. In this paper, we consider $C=69$ atmospheric variables, which will be introduced in detail in Table \ref{Atmospheric-Variables}. To align the experimental setup with ML models like Pangu-Weather~\cite{bi2023accurate} and GraphCast~\cite{lam2023learning}, we adopt a latitude and longitude resolution of $0.25^{\circ }$, which means the global atmospheric data is projected into a two-dimensional plane that $H=721$ and $W=1440$ horizontally.

The model learns to map from $\mathcal X^i$ to $\mathcal X^{i+1}$ with a time interval of 6 hours. Formally, it can be written as:
\begin{equation}
\label{equ:1}
F_\theta(\mathcal X^{i})=P(\mathcal X^{i+1}|\mathcal X^{i})
\end{equation}
where $\theta$ denotes the parameters of the model. By iterating the above process $n$ times, i.e. autoregressive prediction, the model can forecast any future time step $\mathcal X^{i+n}(n\in \mathbb{Z} )$.

\subsection{Model Framework Overview}

\begin{figure}[ht]
% \vspace{-1.0em}
\centerline
{\includegraphics[width=8cm]{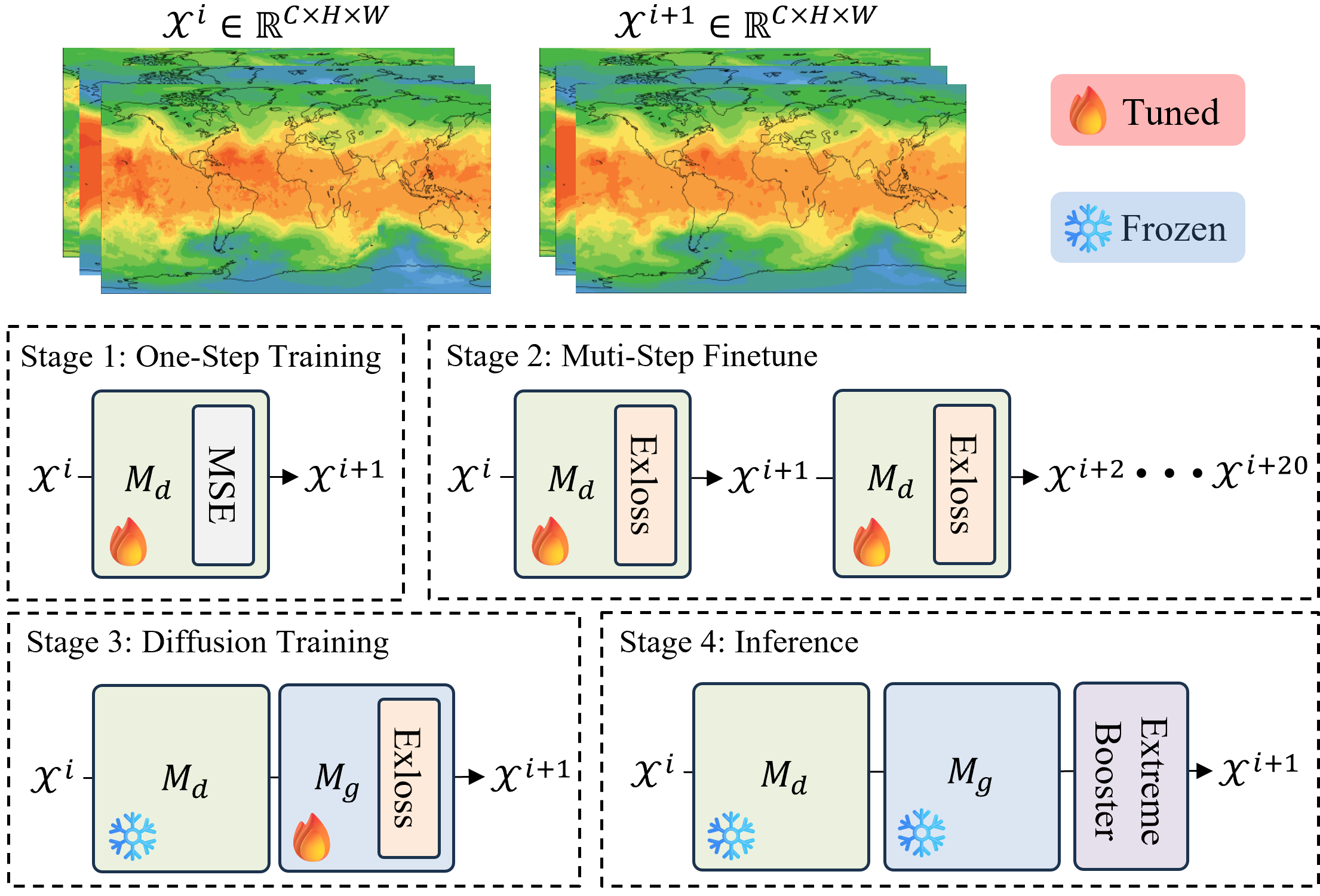}}
\vspace{-0.6em}
\caption{\textbf{Model Framework.} The model consists of three cascaded parts, namely the deterministic forecast model $M_d$, the generation model $M_g$ for enhancing extreme values, and the ExBooster module for modeling uncertainty.}
\label{fig: Framework}
\vspace{-0.8em}
\end{figure}

Our model consists of two trainable modules and one training-free module. They are, respectively, a deterministic model $M_d$, a probabilistic generation model $M_g$, and the ExBooster module. Specifically, $M_d$ learns the mapping from $\mathcal X^{i}$ to $\mathcal X^{i+1}$, followed by $M_g$, a conditional diffusion model that that refines the predictions of $M_d$ with additional details. Ultimately, ExBooster simulates atmospheric uncertainty through random sampling from the output of $M_g$ and amalgamating these samples to derive the final prediction.

In addition to one-step ($\mathcal X^{i} \to \mathcal X^{i+1}$) training, we also incorporate a finetuning of autoregressive regime to improve model's performance in multi-step prediction. The training process of our model can be divided into four stages:

% \begin{itemize}
\begin{compactitem}
\item Stage 1: One-Step pretraining. Use MSE loss to train $M_d$ learning mapping from $\mathcal X^{i}$ to $\mathcal X^{i+1}$.
\item Stage 2: Muti-Step finetuning. Use the proposed Exloss to finetune $M_d$ learning autoregressive forecasting. This stage improves not only the accuracy of the model's multi-step predictions, but also the ability to predict extreme weather due to the use of Exloss.
\item Stage 3: Diffusion training. Use Exloss to train $M_g$ to refine extreme values while freezing $M_d$'s parameters.
\item Stage 4: Inference. ExBooster that does not require training is added for complete weather forecast, enhancing the forecast hit rate of extreme events.
\end{compactitem}
% \end{itemize}

In the following, we will start by analyzing the inherent limitations of MSE, followed by describing the details of the proposed Exloss and the ExBooster.

\subsection{Why MSE Fails to Predict Extreme}
\label{sec:mse}

We abbreviate Equation \ref{equ:1} as $F_\theta(X)=P(Y|X)$, where $X$ is the input of the model, and $Y$ is the target. According to the previous work~\cite{aravkin2012estimation, kendall2018multi}, when the data is normally distributed $Y \sim \mathcal{N} \left ( \mu ,\sigma^2 \right ) $, using MSE loss to predict $Y$ is equivalent to using maximum likelihood estimation~\cite{jiang2009general} to predict the mean $\mu$ of $Y$ given $X$, as shown in Equation \ref{equ:3}:

\vspace{-1.0em}
\begin{equation}
\resizebox{6.7cm}{!}{$
\begin{aligned}
\underset{\theta}{argmin}\ &MSE(Y, \widetilde{Y}), \mathrm{where} \ \widetilde{Y} \equiv F_{\theta}(X) \\
\Leftrightarrow \underset{\theta}{argmin}\ &-log\left(P\left(Y|\mu,\sigma^2\right)\right), \mathrm{where} \ \mu\equiv F_{\theta}(X) \\
= \underset{\theta}{argmin}\ &-log\left(\frac{1}{\sigma\sqrt{2\pi}}exp\left(-\frac{(\mu-Y)^2}{2\sigma^2}\right)\right)\\
= \underset{\theta}{argmin}\ &\frac{(\mu-Y)^2}{2\sigma^2}+log\left(\sigma\sqrt{2\pi}\right)\\
\end{aligned}
$}
\label{equ:3}
\end{equation}

However, this type of optimization has a significant negative impact on extreme values. Consider the introduction of a new variable, denoted as $Y_{M}=max\{Y_1, Y_2, ..., Y_n\}$, where $Y_i(i=1,2..,n)$ represents different samples. In the context of weather forecasting, assuming that $Y_1, Y_2,...,Y_n$ correspond to the temperatures in different regions, then $Y_{M}$ represents the highest temperature in the world. According to the extreme value theory, $Y_{M}$ obeys the extreme value distribution (more specifically, maximum value distribution), that is, $Y_{M} \sim  G(\widetilde{\mu}, \widetilde{\sigma})$ ~\cite{kotz2000extreme}, whose probability density function is:

\vspace{-1.2em}
\begin{equation}
\label{equ:5}
\resizebox{6.7cm}{!}{$
f(Y_{M})=\frac{1}{\widetilde{\sigma}}exp\left(-\frac{Y_{M}-\widetilde{\mu}}{\widetilde{\sigma}} -exp\left(-\frac{Y_{M}-\widetilde{\mu}}{\widetilde{\sigma} } \right)\right)
$}
\end{equation}

where $\widetilde{\mu}$ and $\widetilde{\sigma}$ are its position and scale parameter respectively. It is worth mentioning that, unlike the normal distribution, the extreme value distribution is asymmetric. When employing maximum likelihood estimation for predicting the extreme value $Y_M$, we can derive the optimization objective through a similar procedure:

\vspace{-1.0em}
\begin{equation}
\label{equ:6}
\resizebox{6.7cm}{!}{$
\begin{aligned}
\underset{\theta}{argmin}\ &-log\left(P\left(Y_{M}|\widetilde{\mu},\widetilde{\sigma}\right)\right)\\
=\underset{\theta}{argmin}\ &\frac{Y_{M}-\widetilde{\mu}}{\widetilde{\sigma}}+exp\left(-\frac{Y_{M}-\widetilde{\mu}}{\widetilde{\sigma}}\right) + log\left(\widetilde{\sigma}\right) \\
where \ \widetilde{\mu}&\equiv max\left\{F_{\theta}(X_1), F_{\theta}(X_2),..., F_{\theta}(X_n) \right\}
\end{aligned}
$}
\end{equation}

Define the optimization function obtained by maximum likelihood estimation of this extreme distribution as $obj_M(\cdot)$:

\vspace{-1.0em}
\begin{equation}
\label{def:obj}
\resizebox{6.7cm}{!}{$
obj_M(\widetilde{\mu})=\frac{Y_{M}-\widetilde{\mu}}{\widetilde{\sigma}}+exp\left(-\frac{Y_{M}-\widetilde{\mu}}{\widetilde{\sigma}}\right) + log\left(\widetilde{\sigma}\right)
$}
\end{equation}

It can be proved (in Appendix \ref{A}) that the optimal solution of Equation \ref{equ:6}, that is, the minimum of $obj_M(\cdot)$ is obtained if and only if $\widetilde{\mu}=Y_{M}$, which meets the optimization goal. But due to the asymmetry of the data distribution, $obj_M(\cdot)$ is also asymmetric, as depicted in the upper left corner of Figure \ref{fig: Exloss}, which ultimately results in biased estimates. 

$obj_M(\cdot)$ can be viewed as a loss function that considers the data distribution. Under this assumption, the area enclosed by $obj_M(\cdot)$ and the x-coordinate axis can be regarded as the expectation of the total loss.

As illustrated in Figure \ref{fig: Exloss}, the shading on the left and the shading on the right represent the expectation of the total loss when model provides different estimates (underestimation and overestimation, respectively). Assuming the highest temperature in the world at a certain time is $Y_{M}=40^\circ C$, predicting $\widetilde{\mu}=39.5^\circ C$ will get a lower loss than predicting $\widetilde{\mu}=40.5^\circ C$ based on probability expectation, which encourages the model to predict a smaller $Y_{M}$, that is $\widetilde{\mu}=39.5^\circ C$, rather than $\widetilde{\mu}=40.5^\circ C$. This finally results in underestimating the degree of extremes. The same situation occurs when predicting minimum values, such as the global minimum temperature. For a comprehensive analysis of $obj_M(\cdot)$, please refer to Appendix \ref{A}.

In practice, the highest temperature predicted by the MSE-based model is always smaller than the true value ~\cite{ben2023rise}, which supports the above proof. 

\subsection{Exloss}
Section \ref{sec:mse} proves from EVT that using MSE loss will underestimate the degree of extremes. To address this issue, a simple and straightforward idea is to make the originally asymmetric $obj_M(\cdot)$ symmetrical through scaling. 

It is cumbersome and unnecessary to transform $obj_M(\cdot)$ into a completely symmetric function. In actual operation, we use linear scaling to make the optimization objective function $obj_M(\cdot)$ symmetrical within a certain range $\epsilon$, where $\epsilon$ is a hyperparameter that depends on both the data distribution and the accuracy of the model's predictions. In simple terms, when the model's predictions are highly accurate, the prediction error is smaller, allowing for the selection of a smaller value for $\epsilon$. For detailed guidelines on selecting an appropriate value for $\epsilon$, please refer to Appendix \ref{B}. Mathematically, the process of achieving symmetry through scaling can be expressed as follows:

% \vspace{-1.0em}

\begin{equation}
\label{equ:7.1}
\resizebox{7.5cm}{!}{$
\begin{aligned}
&\widetilde{\mu}_s=\left\{\begin{matrix}
\frac{\widetilde{\sigma}}{s_1} (\widetilde{\mu}-Y_M)+Y_M,\ \widetilde{\mu}\le Y_M \\
\frac{\widetilde{\sigma}}{s_2} (\widetilde{\mu}-Y_M)+Y_M,\ \widetilde{\mu}>  Y_M  \\
\end{matrix}\right. \\
&s.t. \int_{\widetilde{\mu}_s=Y_{M}-\epsilon}^{Y_{M}} obj_M\left(\widetilde{\mu}_s\right ) \ \mathrm{d}\widetilde{\mu}_s  = \int_{\widetilde{\mu}_s=Y_{M}}^{Y_{M}+ \epsilon } obj_M(\widetilde{\mu}_s)\ \mathrm{d}\widetilde{\mu}_s\\
\end{aligned}
$}
\end{equation}

The purpose of the scaling factors $s_1$ and $s_2$ is to equalize the areas of the two shaded regions in Figure \ref{fig: Exloss}, which ensures that the expectations of the total loss are the same for underestimation and overestimation. Subsequently, $s_1$ and $s_2$ are combined into a function $S(\hat{\mathcal Y}, \mathcal Y)$. By applying scaling function to the loss, we propose Exloss:

\vspace{-1.5em}

\begin{equation}
\label{equ:7.2}
\resizebox{6.7cm}{!}{$
\begin{aligned}
Exloss\left (\hat{\mathcal Y},\mathcal Y\right )&=\left \|  S(\hat{\mathcal Y},\mathcal Y)\odot \hat{\mathcal Y}- S(\hat{\mathcal Y},\mathcal Y)\odot  \mathcal Y\right \| ^2 \\
&=\left \| S(\hat{\mathcal Y},\mathcal Y)\odot \left(\hat{\mathcal Y}- \mathcal Y\right) \right \|^2
\end{aligned}
$}
\end{equation}
where $\hat{\mathcal Y}$ and $\mathcal Y$ represent the model prediction and the target respectively, $S(\hat{\mathcal Y},\mathcal Y) \in \mathbb{R} ^{C\times H \times W}$ is \textbf{asymmetrical} scaling function. For the detailed form of $S(\hat{\mathcal Y}, \mathcal Y)$, please refer to Appendix \ref{B}.

In fact, by calculating $S(\hat{\mathcal{Y}}, \mathcal{Y})$, it can be observed that Exloss increases the loss weight when the model underestimates extreme value. This behavior is highly rational, as it adaptively increases the penalty for underestimations.

\subsection{ExBooster}

The atmospheric system has uncertainty~\cite{thompson1957uncertainty}, which means that for a given initial condition, there are multiple possible predictions. Some low-probability extreme events only exist in some prediction results. In such scenarios, multiple sampling of predictions aids in capturing a broader spectrum of potential extreme events. 

In this work, we propose the ExBooster, drawing insights from ensemble algorithms~\cite{gneiting2005weather}. The process begins by introducing noise to the prediction, generating $m$ ensemble results. Due to the addition of noise, these different prediction results form a larger range of values than the original single prediction, which means that they contain more extreme events. Then, we use the ranking histogram method~\cite{hamill2001interpretation} to integrate these m prediction results. Specifically, all pixels in these samples are first sorted and grouped by the size of $m$, and then the original value is replaced with the median of the corresponding part according to the ranking of each element in the initial prediction. 

\begin{figure}[ht]
\vspace{-0.6em}
\centerline
{\includegraphics[width=8.5cm]{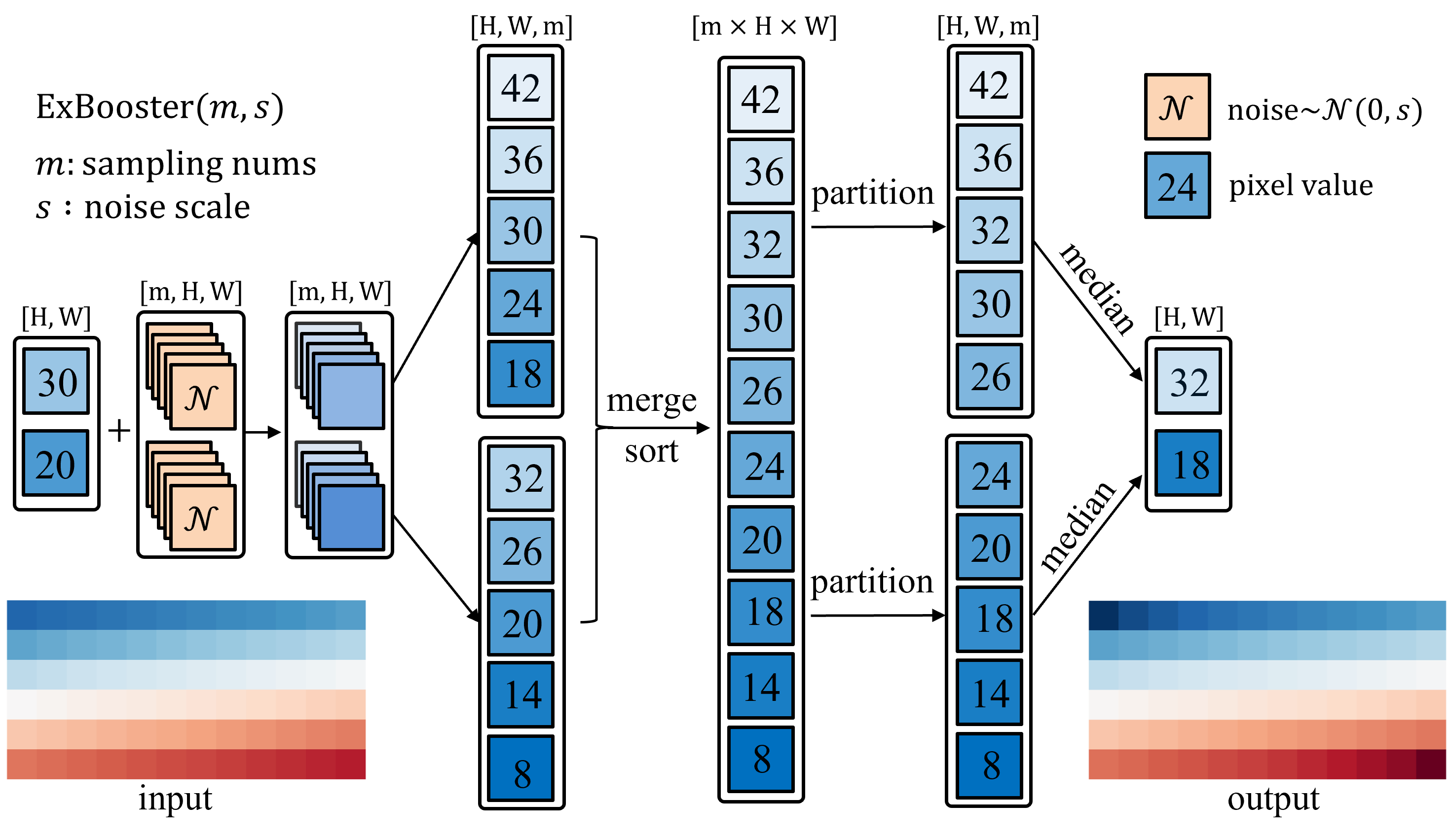}}
\vspace{-0.6em}
\caption{\textbf{ExBooster.} Multiple random samplings can simulate forecast uncertainty and enhance the accuracy of extreme event predictions. The visual example demonstrates that extreme values are more pronounced in the output.}
\label{fig:ens}
\vspace{-0.2em}
\end{figure}

As depicted in Figure \ref{fig:ens}, ExBooster adaptively expands the range of pixel values without altering the overall distribution, thereby producing more extreme predictions. In practice, we choose $m=50$, that is, randomly sample $50$ predictions, which is enough to mimic the uncertainty of atmospheric dynamics~\cite{price2023gencast}. For detailed implementation, please refer to Appendix \ref{C}.

\section{Experiment}

\subsection{Experimental Setup}
\paragraph{Dataset.} The training process utilizes the 5th generation of ECMWF reanalysis (ERA5) data ~\cite{hersbach2020era5}. We use the data from 1979 to 2017 as the training set, from 2018 to 2021 as the validation set, and from 2022 to 2022 as the test set. Our model employs 69 atmospheric variables, including four surface variables and five upper-level variables with 13 pressure levels, as shown in Table \ref{Atmospheric-Variables}.

\begin{table}[ht]
\vspace{-0.5em}
% \vskip 0.15in
\begin{center}
\resizebox{0.8\columnwidth}{!}{
\begin{tabular}{llll}
\toprule
Name & Description & Levels \\
\midrule
u10    & X-direction wind at 10m height  & Single\\
v10    & Y-direction wind at 10m height  & Single\\
t2m    & Temperature at 2m height        & Single\\
msl    & Mean sea level pressure         & Single\\
z      & Geopotential                    & 13\\
q      & Absolute humidity               & 13\\
u      & X-direction wind                & 13\\
v      & Y-direction wind                & 13\\
t      & Temperature                     & 13\\
\bottomrule
\end{tabular}
}
\end{center}
% \vskip -0.1in
\vspace{-0.8em}
\caption{\textbf{Atmospheric Variables.} The levels are 50,100,150, 200,250,300,400,500,600,700,850,925,1000hPa.}
\label{Atmospheric-Variables}
\end{table}
\vspace{-1.0em}

\paragraph{Network Structure.} 
$M_d$ has the same network structure as ~\cite{chen2023fengwu}, which uses Swin-Transformer ~\cite{vaswani2017attention} as backbone, and designs the encoder and decoder through down-sampling and up-sampling. $M_g$ uses a mainstream diffusion implementation which use a U-Transformer ~\cite{petit2021u} as backbone, and realizes conditional generation by concatenating condition to the model input. See Appendix \ref{D} for hyperparameters.

\paragraph{Baseline Models.} 
We choose five ML models, namely Pangu ~\cite{bi2023accurate}, GraphCast ~\cite{lam2023learning}, FengWu ~\cite{chen2023fengwu}, FuXi ~\cite{chen2023fuxi}, FuXi-Extreme ~\cite{zhong2023fuxi} and a physics-based dynamic model ECMWF-IFS ~\cite{model2003ifs}, as the baseline models for comparative experiments. For detailed introductions of these models, please refer to the Related Work.

\paragraph{Metrics.}
We use three metrics to reflect the extreme weather forecast performance and the overall accuracy.

% \begin{itemize}
\begin{compactitem}
\item \textbf{RQE.} Relative Quantile Error ~\cite{kurth2023fourcastnet}. It is defined as the relative error of quantiles (90\%-99.99\%) between prediction and target. A positive RQE value indicates that extreme values are overestimated, while a negative value indicates underestimation.
\item \textbf{SEDI.} Symmetric Extremal Dependency Index ~\cite{ferro2011extremal}. By classifying each pixel into extreme weather or normal weather, this metric calculates the hit rate of the classification. By selecting thresholds, SEDI for different degrees of extreme weather can be calculated, such as SEDI 90th, SEDI 95th, etc ~\cite{zhong2023fuxi, han2024cra5}. A higher SEDI value closer to 1 indicates better prediction accuracy for extreme values.
\item \textbf{RMSE.} Weighted Root Mean Square Error ~\cite{rasp2020weatherbench}. It reflects the overall forecast error. A smaller RMSE value indicates a higher forecast accuracy.
\end{compactitem}
% \end{itemize}

Both RQE and SEDI serve as metrics to evaluate the accuracy of extreme value predictions, but their focus is different. RQE reflects more on the global scale, while SEDI reflects more on the regional (pixel-level) scale.

\begin{figure*}[htbp]
\centerline
{\includegraphics[width=16cm]{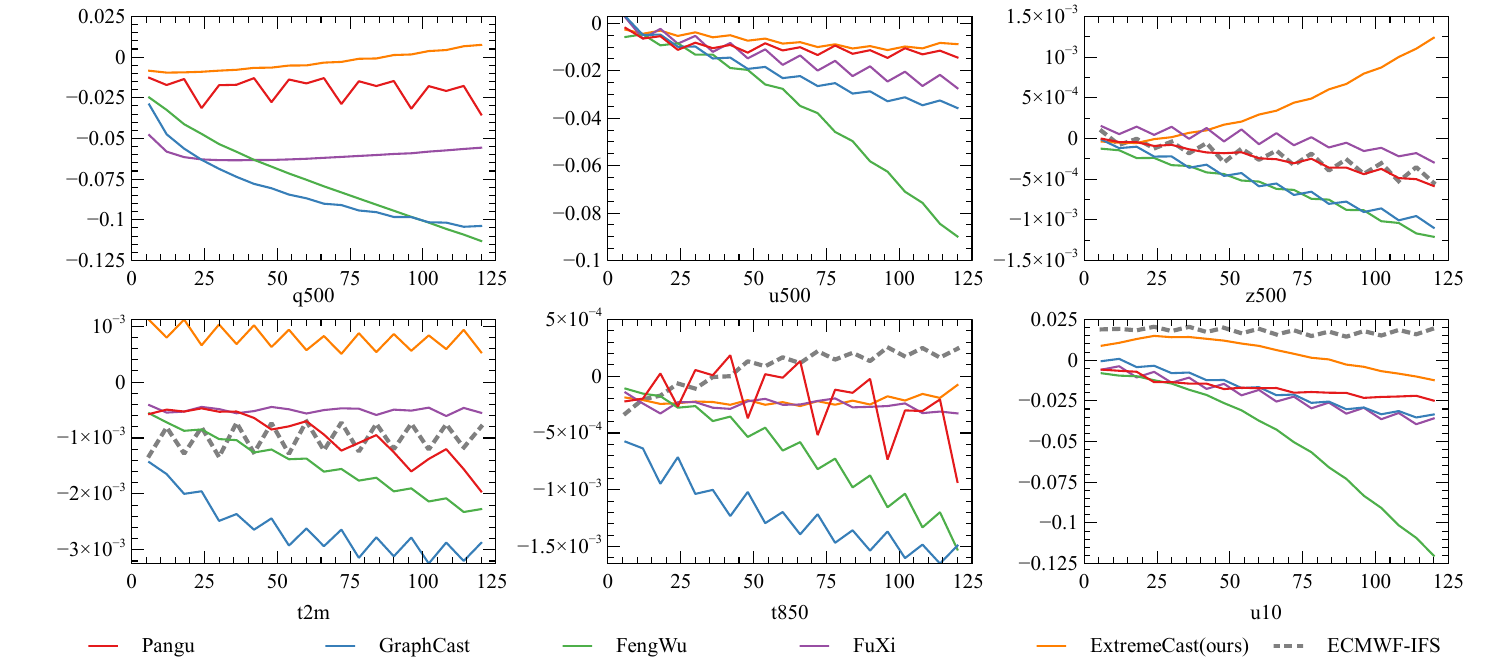}}
\caption{\textbf{RQE} ($RQE<0$ means underestimating extreme values, $RQE>0$ means overestimating extreme values). All results are tested on 2018 data and use ERA5 as the target.}
\label{fig: rqe}
\end{figure*}

\begin{figure*}[htbp]
\centerline
{\includegraphics[width=18cm]{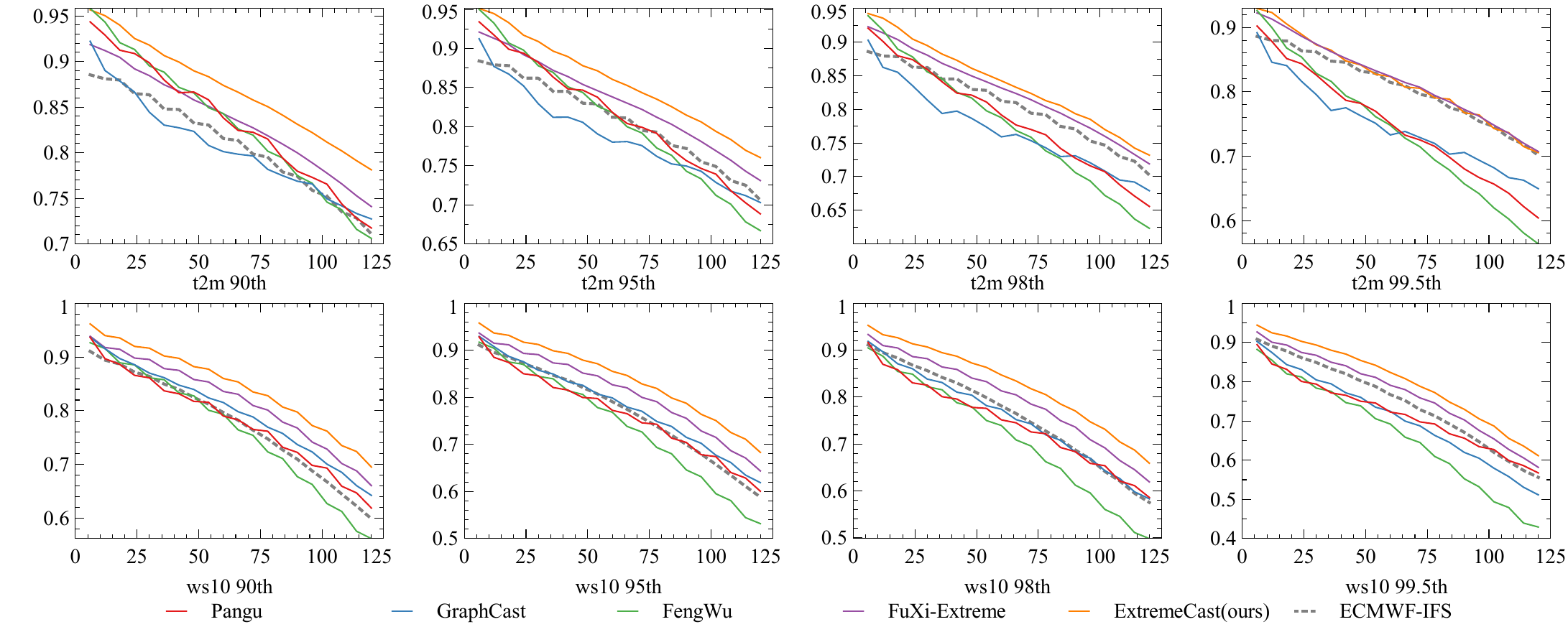}}
\caption{\textbf{SEDI} (the closer to 1 the better). ws10 represents surface wind speed, that is, $ws10=\sqrt{u10^2+v10^2}$. All results are tested on 2018 data and use ERA5 as the target.}
\label{fig: sedi}
\end{figure*}

\begin{figure*}[htbp]
\centerline
{\includegraphics[width=18cm]{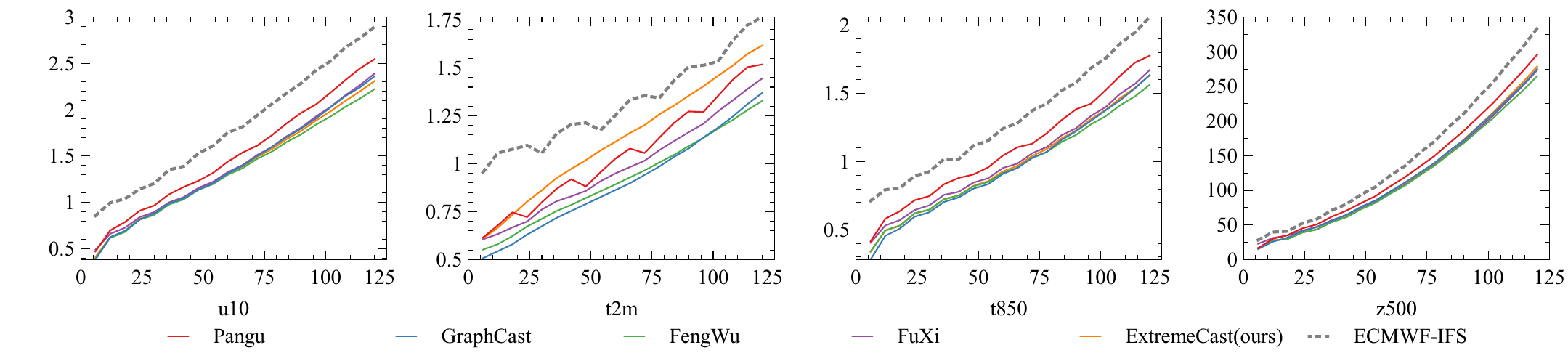}}
\caption{\textbf{RMSE} (the smaller the better). All results are tested on 2018 data and use ERA5 as the target.}
\label{fig: wrmse}
\end{figure*}

\begin{figure*}[ht]
\vspace{-0.4em}
\centerline
{\includegraphics[width=17.0cm]{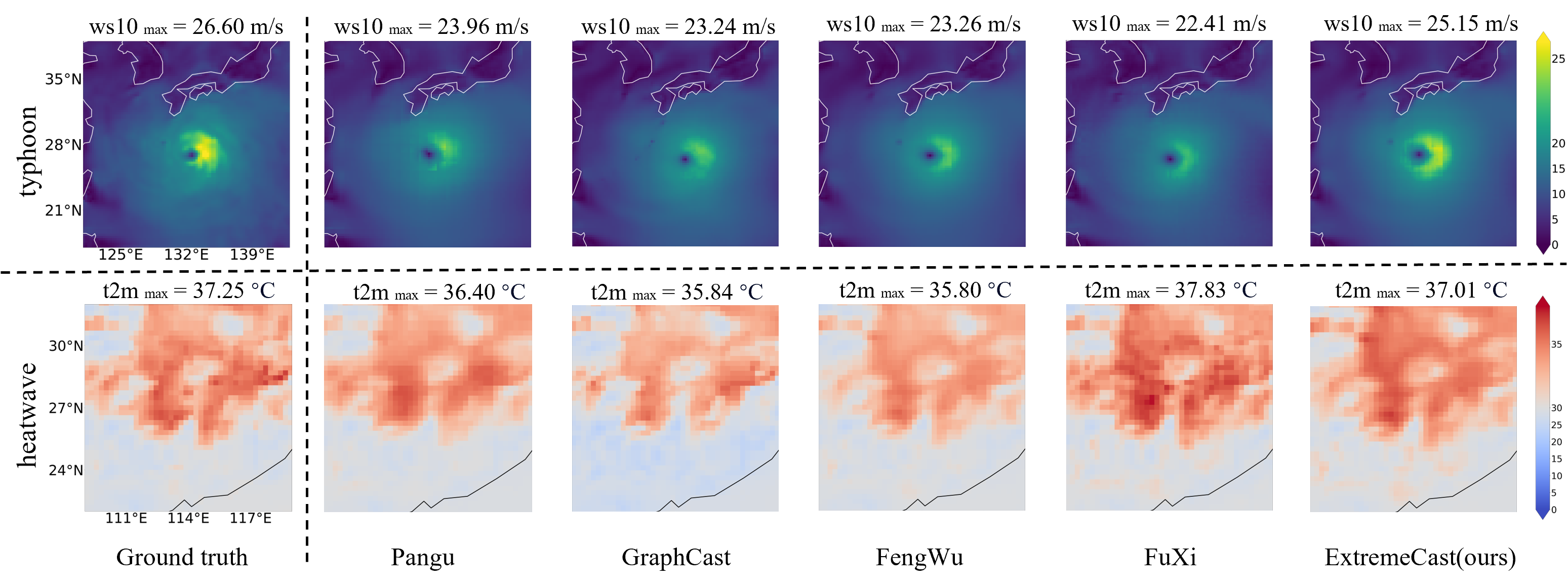}}
\vspace{-1.0em}
\caption{\textbf{Three-day Typhoon and Heatwave Forecast Visualization.} The model input and ground truth are from ERA5.}
\label{fig: typhoon}
\end{figure*}

\begin{table*}
    \vspace{-0.4em}
    \centering
    \resizebox{16cm}{!}{
    \begin{tabular}{cccc|p{1.5cm}p{1.5cm}p{1.5cm}p{1.5cm}|p{1.5cm}p{1.5cm}p{1.5cm}p{2cm}}
        \toprule
                &      &   &   & \multicolumn{4}{c|}{RQE (- underestimate, + overestimate)} & \multicolumn{4}{c}{SEDI (the closer to 1 the better)} \\
        &Exloss  & Diffusion & ExBooster & t2m & u10 & q500 & u500 & t2m@90th & t2m@99.5th	& ws10@90th & ws10@99.5th \\
        \hline
        
        \ding{192}&\texttimes & \texttimes & \texttimes & -0.0023 & -0.1201 & -0.1130 & -0.0899 & 0.7059 & 0.5637 & 0.5623 & 0.4295\\

        \hdashline
        \ding{193}&\checkmark & \texttimes & \texttimes & -0.0012 & -0.0792 & -0.0330 & -0.0538 & 0.7589 & 0.6222 & 0.6423 & 0.5156\\

        \hdashline[1pt/4pt]
        \ding{194}&\texttimes & \checkmark & \checkmark & +0.0005 & -0.0296 & -0.0407 & -0.0299 & 0.7743 & 0.6947 & 0.6793 & 0.5778\\

        \ding{195}&\checkmark & \texttimes & \checkmark & -0.0007 & -0.0374 & \textbf{+0.0063} & -0.0173 & 0.7643 & 0.6405 & 0.6791 & 0.5773\\
        
        \ding{196}&\checkmark & \checkmark & \texttimes & \textbf{+0.0002} & -0.0496 & -0.0291 & -0.0459 & 0.7764 & 0.6927 & 0.6627 & 0.5517\\

        \hdashline
        \ding{197}&\checkmark & \checkmark & \checkmark & +0.0008 & \textbf{-0.0120} & +0.0092 & \textbf{-0.0125} & \textbf{0.7796} & \textbf{0.7106} & \textbf{0.6930} & \textbf{0.6056}\\
        \bottomrule
    \end{tabular}
    }
    \vspace{-0.4em}
    \caption{\textbf{Ablation Experiment.} All results are predictions on the fifth day and use ERA5 as the target.}
    \label{tab:Ablation}
    \vspace{-1.2em}
\end{table*}

\subsection{Global Extreme Weather Forecast Capability}
We use the RQE to evaluate the models' ability to predict extreme values on the global scale. As depicted in Figure \ref{fig: rqe}, the RQE values of most ML models exhibit a tendency to become progressively more negative as the lead time increases, indicating an increasing underestimation of extreme values. On the contrary, the dynamic model has a relatively better RQE, that is, an RQE value closer to 0, and does not decrease as the lead time increases.

For our model, overall, the RQE does not decrease as the lead time increases, which is due to Exloss's asymmetric optimization, avoiding the underestimation of extreme values. Especially for q500 (the channel of absolute humidity at the 500hPa pressure layer), u500, and u10, our model predicts with a RQE close to 0 consistently, surpassing both ML models and the dynamic model, showing the accurate and stable extreme value prediction capabilities. 

Our model slightly overestimates (around $10^{-3}$) the extreme values in certain atmospheric variables like z500, which is very rare for ML models but is occasionally observed in dynamic models as well. Appropriate overestimation can provide a more adequate upper bound for the prediction of extreme events, which is beneficial for some applications like disaster early warning ~\cite{li2021evaluating}.

\subsection{Regional Extreme Weather Forecast Capability}

Different from RQE, SEDI focus more on extreme values in local areas, because each pixel uses its quantile in the time dimension as a threshold that varies from month to month. For instance, a temperature of $-10^\circ C$ may not be considered as an extreme event for high-latitude areas during winter, but it could be extreme for mid-latitude areas in summer.

As illustrated in Figure \ref{fig: sedi}, for surface temperature, our model's prediction accuracies at 90th, 95th, 98th extreme degrees are significantly higher than that of other models. Regarding the most extreme scenarios at the 99.5th quantile, our model demonstrates relatively better performance when the lead time is less than 24 hours, and for lead times exceeding 24 hours, our model has similar performance to FuXi-Extreme and ECMWF-IFS.

For surface wind speed, our model significantly outperforms other models across all extreme levels. It is worth noting that the SEDI of our model's first step prediction (lead time=6h) is remarkably high, predominantly exceeding 0.95. This is particularly challenging for wind speed which exhibit rapid and diverse changes, highlighting the excellent performance of our model in forecasting extreme wind speeds.

\subsection{Case Analysis}
To better showcase the performance of our model, we select two natural disasters, typhoon and heatwave, for visual analysis. Specifically, we choose Nanmadol typhoon ~\cite{wu2023interaction} at 6:00 on September 17, 2022, and the heatwave in the southeastern coastal area of China at 12:00 on August 14, 2022 ~\cite{jiang2023extreme}, for three-day forecasting. Figure \ref{fig: typhoon} illustrates the results, with the two rows showing surface wind speed and surface temperature, respectively.

Regarding typhoons, the wind speeds of our prediction are the most extreme and closest to the real wind speeds. This aligns with the findings observed in the far-leading SEDI of wind speed, showing the exceptional performance of our model in predicting extreme wind speeds. This capability is of significant importance for estimating the intensity of typhoon landings and assessing the potential damage to coastal areas.

In the case of the heatwave, both of the predictions of our model and FuXi-Extreme are more extreme than other models. Although fuxi-extreme's predictions are more extreme than ours, our model gets a prediction ($37.01^\circ C$) that is closer to the true value ($37.25^\circ C$) for the regional maximum temperature prediction. For more visualizations, please refer to Appendix \ref{E}.

\subsection{Ablation Experiment}

We assess the individual contributions of each module by conducting ablation experiments where we remove Exloss, and ExBooster. Additionally, as an effective implementation component, diffusion model is also considered in the ablation. 

Through Table \ref{tab:Ablation}, we find that all three modules are beneficial in enhancing the prediction of extreme values. Specifically, we discover that the asymmetric optimization of Exloss plays a crucial role in enhancing the forecasting capabilities of extreme weather events. This is evident from the comparison between row \ding{192} and row \ding{193}, where the use of Exloss leads to significant improvements in both RQE and SEDI metrics. Furthermore, comparing row \ding{194} and row \ding{197} (best version), the lack of Exloss results in a significant decrease in all metrics.

Moreover, Diffusion and ExBooster have great improvements in the prediction of extreme temperatures and extreme wind speeds respectively, which highlights the favorable decoupling properties of different modules in our model.

\subsection{Normal Weather Forecast Capability}
Although the focus of this paper is extreme weather prediction, the overall forecast accuracy of our model is still competitive. As shown in Figure \ref{fig: wrmse},  while our model demonstrates outstanding performance in predicting extreme wind speeds, it also achieves a high accuracy in overall wind speed prediction, as indicated by RMSE u10. Additionally, the RMSE metrics of our model on other variables are also comparable to top ML models and surpass ECMWF-IFS.

\section{Conclusions}

Weather forecast models based on machine learning often face challenges in accurately predicting extreme weather. In this paper, we provide an explanation based on extreme value theory, for the inherent limitations of MSE loss in forecasting extreme values. To address this issue, a novel loss function, Exloss, is introduced, which rectifies deviations in extreme value prediction through asymmetric optimization. 

Moreover, we propose a training-free module named ExBooster that models the uncertainty of atmosphere through random sampling and integrates multiple forecast results to improve the forecast hit rate of extreme events. Experimental results show that our model, ExtremeCast, achieves SOTA performance in extreme value metrics such as RQE and SEDI, while maintaining the competitive overall accuracy.

\textbf{The limitation of our model} lies in the increased time cost resulting from the inclusion of the cascaded pipeline, and the sensitivity of ExBooster affected by the scale of noise.

\bibliography{aaai25}

\clearpage
\appendix
\onecolumn
\section{Analysis of the Optimization Objective}
\label{A}
First, we examine the optimization function for the scenario where the model predicts the maximum value $Y_{M}=\max\{Y_1, Y_2, ..., Y_n\}$. In this case, $Y_{M}$ follows a maximum distribution, characterized by its probability density function ~\cite{wellander2001maximum}:

\begin{equation}
\label{equ:a1}
f(Y_{M})=\frac{1}{\widetilde{\sigma}}exp\left(-\frac{Y_{M}-\widetilde{\mu}}{\widetilde{\sigma}} -exp\left(-\frac{Y_{M}-\widetilde{\mu}}{\widetilde{\sigma} } \right)\right)
\end{equation}

where $\widetilde{\mu}$ and $\widetilde{\sigma}$ are its position and scale parameter respectively. Unlike the normal distribution, maximum distribution is an asymmetric skewed distribution. We can draw a histogram of the maximum distribution through the following algorithm, so as to observe the characteristics of the maximum distribution more directly.

\begin{algorithm}[ht]
   \caption{Maximum Distribution Sample}
   \label{alg:MaximumDistributionSample}
\begin{algorithmic}
   \STATE $M=[\ ] \  \# \   $ List of maximum distribution sampling
   \FOR{$i=1$ {\bfseries to} $10000$}
       \STATE $N=[\ ]\  \# \   $ List of normal distribution sampling
       \FOR{$j=1$ {\bfseries to} $10000$}
            \STATE
            $x\longleftarrow \mathcal{N}(0, 1)$
            \STATE
            $N$\bfseries .append($x$)
       \ENDFOR
       \STATE
        $x_{M}\longleftarrow $ \bfseries Max($N$)
        \STATE
        $M$\bfseries .append($x_{M}$)
   \ENDFOR
   \STATE
   Draw a histogram of $M$ and $N$.
\end{algorithmic}
\end{algorithm}

\begin{figure}[ht]
\centerline
{\includegraphics[width=12cm]{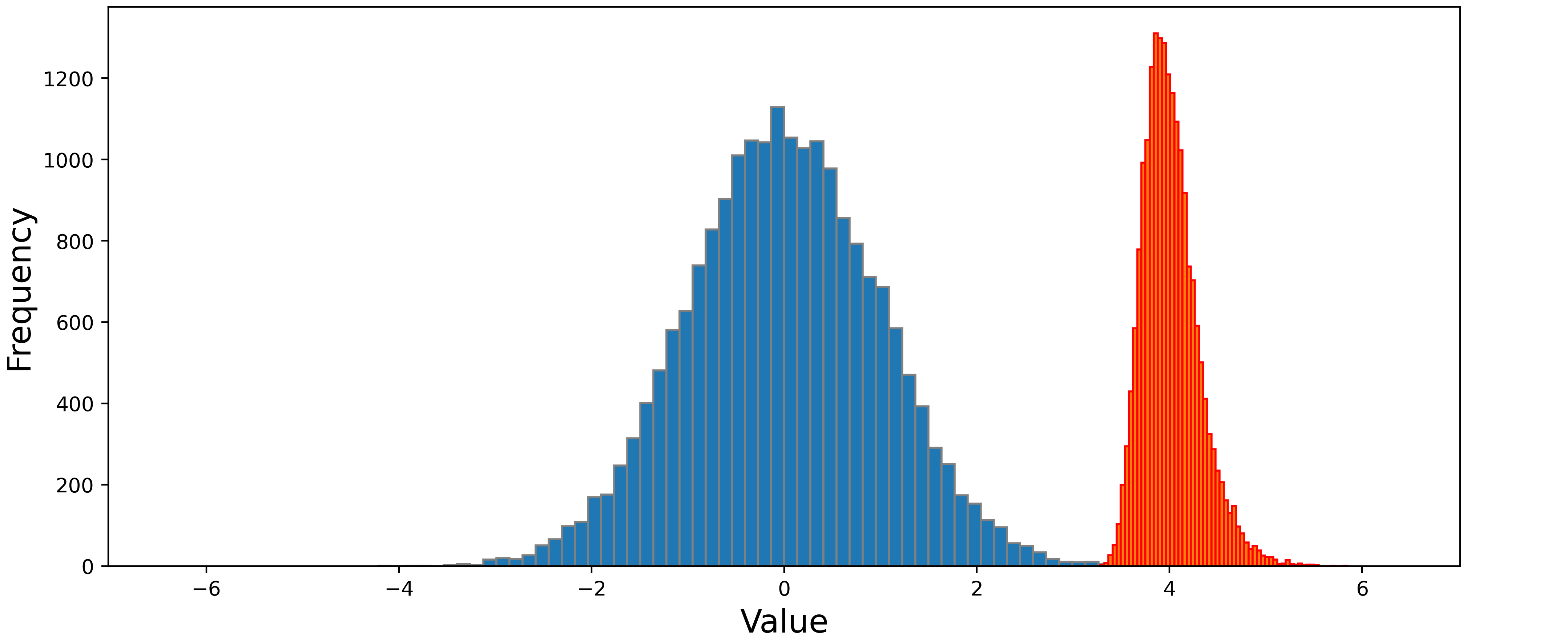}}
\caption{\textbf{Visualization of Maximum Distribution and Normal Distribution.} The blue histogram is sampled from the normal distribution, and the red histogram is sampled from the maximum distribution.}
\label{fig: max_dist}
\end{figure}

The asymmetry of the distribution of maximum values can be clearly seen from Figure \ref{fig: max_dist}. This asymmetry is also the focus of our discussion below.

When making predictions for the extreme value $Y_M$, the optimization objective can be formulated as follows:

\begin{equation}
\label{equ:a2}
\begin{aligned}
\underset{\theta}{argmin}\ &-log\left(P\left(Y_{M}|\widetilde{\mu},\widetilde{\sigma}\right)\right)\\
=\underset{\theta}{argmin}\ &\frac{Y_{M}-\widetilde{\mu}}{\widetilde{\sigma}}+exp\left(-\frac{Y_{M}-\widetilde{\mu}}{\widetilde{\sigma}}\right) + log\left(\widetilde{\sigma}\right) \\
where \ \widetilde{\mu}&\equiv max\left\{F_{\theta}(X_1), F_{\theta}(X_2),..., F_{\theta}(X_n) \right\}
\end{aligned}
\end{equation}

Denote the optimization objective as $obj_M(\cdot)$:

\begin{equation}
\label{equ:a3}
obj_M(\widetilde{\mu})=\frac{Y_{M}-\widetilde{\mu}}{\widetilde{\sigma}}+exp\left(-\frac{Y_{M}-\widetilde{\mu}}{\widetilde{\sigma}}\right) + log\left(\widetilde{\sigma}\right)
\end{equation}

The derivative of $obj_M(\cdot)$ with respect to $\widetilde\mu$ is:

\begin{equation}
\label{equ:a4}
obj_M'(\widetilde{\mu}) =-\frac{1}{\widetilde{\sigma}}+\frac{1}{\widetilde{\sigma}}exp\left(-\frac{Y_{M}-\widetilde{\mu}}{\widetilde{\sigma}}\right) 
\end{equation}

Its second derivative is:
\begin{equation}
\label{equ:a5}
obj_M''(\widetilde{\mu}) =\frac{1}{\widetilde{\sigma}^2}exp\left(-\frac{Y_{M}-\widetilde{\mu}}{\widetilde{\sigma}}\right) 
\end{equation}

Its second derivative $obj_M''(\widetilde{\mu})$ is positive, indicating that the derivative, $obj_M'(\widetilde{\mu})$, monotonically increases. As $obj_M'(Y_M)=0$, we can conclude that:

\begin{equation}
\label{equ:a6}
\left\{\begin{matrix}
obj_M'(\widetilde{\mu})<0,\widetilde{\mu}<Y_M\\
obj_M'(\widetilde{\mu})=0,\widetilde{\mu}=Y_M\\
obj_M'(\widetilde{\mu})>0,\widetilde{\mu}>Y_M
\end{matrix}\right.
\end{equation}

So $obj_M(\widetilde{\mu})$ first decreases and then increases, with its minimum value being:

\begin{equation}
\label{equ:a7}
min(obj_M(\widetilde{\mu}))=obj_M(\widetilde{\mu})|_{\widetilde{\mu}=Y_M}=1+log\left(\widetilde{\sigma}\right)
\end{equation}

Next, we examine the situation where the model predicts the minimum value $Y_{m}=\min\{Y_1, Y_2, ..., Y_n\}$. In this case, $Y_{m}$ follows a minimum distribution, whose probability density function is ~\cite{cai2010minimum}:

\begin{equation}
\label{equ:a8}
f(Y_{m})=\frac{1}{\widetilde{\sigma}}exp\left(\frac{Y_{m}-\widetilde{\mu}}{\widetilde{\sigma}} -exp\left(\frac{Y_{m}-\widetilde{\mu}}{\widetilde{\sigma} } \right)\right)
\end{equation}

where $\widetilde{\mu}$ and $\widetilde{\sigma}$ are its position and scale parameter respectively. When making predictions for the extreme value $Y_m$, the optimization objective is:

\begin{equation}
\label{equ:a9}
\begin{aligned}
\underset{\theta}{argmin}\ &-log\left(P\left(Y_{m}|\widetilde{\mu},\widetilde{\sigma}\right)\right)\\
=\underset{\theta}{argmin}\ &-\frac{Y_{m}-\widetilde{\mu}}{\widetilde{\sigma}}+exp\left(\frac{Y_{m}-\widetilde{\mu}}{\widetilde{\sigma}}\right) + log\left(\widetilde{\sigma}\right) \\
where \ \widetilde{\mu}&\equiv min\left\{F_{\theta}(X_1), F_{\theta}(X_2),..., F_{\theta}(X_n) \right\}
\end{aligned}
\end{equation}

Denote the optimization objective as $obj_m(\cdot)$

\begin{equation}
\label{equ:a10}
obj_m(\widetilde{\mu})=-\frac{Y_{m}-\widetilde{\mu}}{\widetilde{\sigma}}+exp\left(\frac{Y_{m}-\widetilde{\mu}}{\widetilde{\sigma}}\right) + log\left(\widetilde{\sigma}\right)
\end{equation}

Similar to the analysis process of $obj_M(\cdot)$, we can prove that $obj_m(\cdot)$ first decreases and then increases. The minimum value of $obj_m(\cdot)$ is:

\begin{equation}
\label{equ:a11}
min(obj_m(\widetilde{\mu}))=obj_m(\widetilde{\mu})|_{\widetilde{\mu}=Y_m}=1+log\left(\widetilde{\sigma}\right)
\end{equation}

\begin{figure}[ht]
\centerline
{\includegraphics[width=16cm]{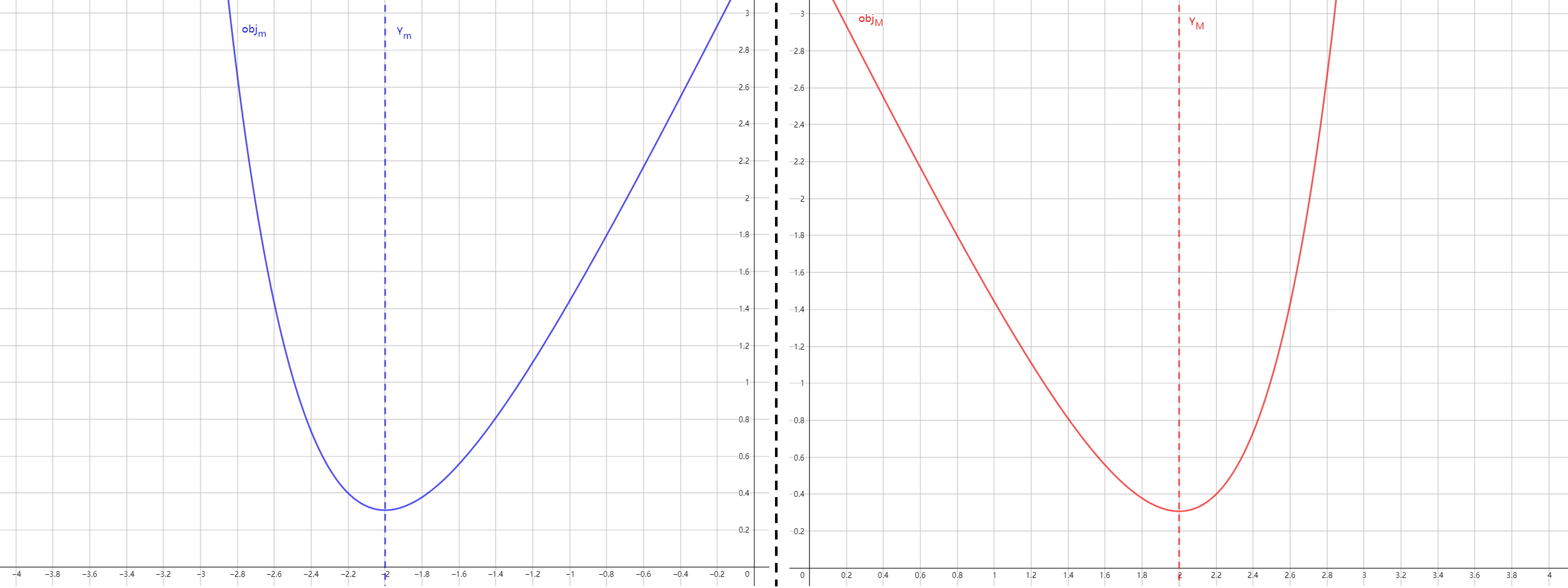}}
\caption{\textbf{Visualization of Optimization Functions.} The left figure is $obj_m(\cdot)$ when $Y_m=-2$ and $\sigma=1/2$. The right figure is $obj_M(\cdot)$ when $Y_M=2$ and $\sigma=1/2$. The x-coordinate represents $\widetilde{\mu}$ predicted by the model.}
\label{fig: min_max}
\end{figure}

By visualizing the shapes of $obj_M(\cdot)$ and $obj_m(\cdot)$, we gain a clearer insight into their respective characteristics. It can be seen from Figure \ref{fig: min_max} that both $obj_m(\cdot)$ and $obj_M(\cdot)$ are asymmetric.

\section{Scaling Function}
\label{B}
$obj_M(\cdot)$ and $obj_m(\cdot)$ can actually be regarded as loss functions that take into account the distribution characteristics of the original data ~\cite{wang2020comprehensive}. Under this assumption, the area enclosed by the function and the x-coordinate axis can be regarded as the expectation of the total loss of this part of the sample ~\cite{hernandez2012unified}.

We hope that the areas of $obj_M(\cdot)$ near $Y_M$ are equal, thereby ensuring that the expectations of the total loss when underestimating extreme values and overestimating extreme values are equal. The same goes for $obj_m(\cdot)$. This goal is accomplished through scaling. Formally, for $obj_M(\cdot)$, it can be written:

\begin{equation}
\label{equ:a12}
\begin{aligned}
&\widetilde{\mu}_s=\left\{\begin{matrix}
\frac{\widetilde{\sigma}}{s_1} (\widetilde{\mu}-Y_M)+Y_M,\ \widetilde{\mu}\le Y_M \\
\frac{\widetilde{\sigma}}{s_2} (\widetilde{\mu}-Y_M)+Y_M,\ \widetilde{\mu}>  Y_M  \\
\end{matrix}\right. \\
&s.t. \int_{\widetilde{\mu}_s=Y_{M}-\epsilon}^{Y_{M}} obj_M\left(\widetilde{\mu}_s\right ) \ \mathrm{d}\widetilde{\mu}_s  = \int_{\widetilde{\mu}_s=Y_{M}}^{Y_{M}+ \epsilon } obj_M(\widetilde{\mu}_s)\ \mathrm{d}\widetilde{\mu}_s\\
\end{aligned}
\end{equation}

To simplify the calculation, we subtract the minimum value of $obj_M(\cdot)$, which is $1+\log(\widetilde{\sigma})$, from $obj_M(\cdot)$. This adjustment ensures that the lowest value of $obj_M\cdot$ becomes 0, as shown in Figure \ref{fig: r1_r2}. We can then compute the area around the lowest point by performing integration:

\begin{figure}[ht]
\centerline
{\includegraphics[width=10cm]{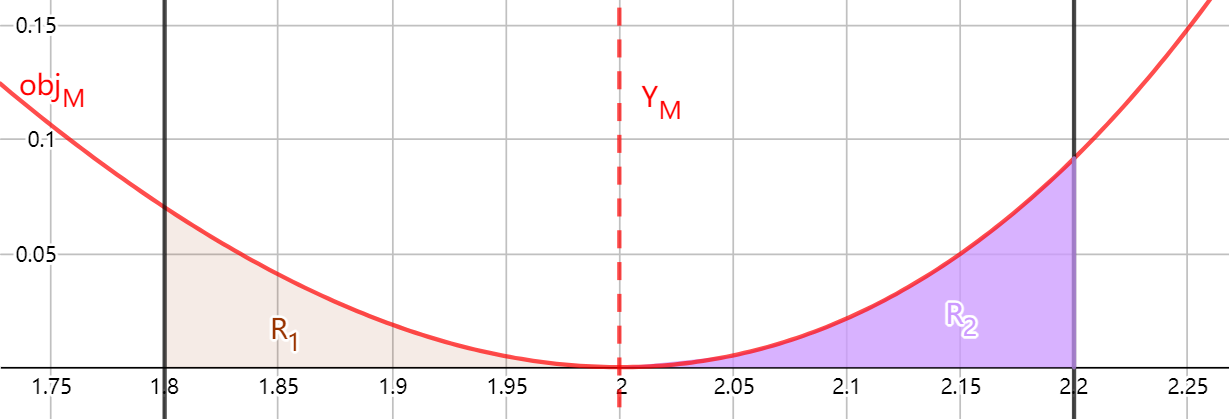}}
\caption{\textbf{Area Integral.} We hope that by scaling, the area of $R_1$ will be the same as $R_2$.}
\label{fig: r1_r2}
\end{figure}

The area integral of $R_1$ is as follows:

\begin{equation}
\label{equ:a13}
\begin{aligned}
&\int_{\widetilde{\mu}_s=Y_{M}-\frac{\widetilde{\sigma}}{s_1}\epsilon}^{Y_{M}}obj_M\left(\widetilde{\mu}_s\right ) \ \mathrm{d}\widetilde{\mu}_s,\ \widetilde{\mu}_s=\frac{\widetilde{\sigma}}{s_1}  (\widetilde{\mu}-Y_M)+Y_M \\
=&\int_{\widetilde{\mu}=Y_{M}-\epsilon}^{Y_{M}} \left( \frac{Y_{M}-\left (\frac{\widetilde{\sigma}}{s_1} (\widetilde{\mu}-Y_M)+Y_M\right )}{\widetilde{\sigma}}+exp\left(-\frac{Y_{M}-\left (\frac{\widetilde{\sigma}}{s_1} (\widetilde{\mu}-Y_M)+Y_M\right )}{\widetilde{\sigma}}\right)-1\right)\ \mathrm{d}\left (\frac{\widetilde{\sigma}}{s_1} (\widetilde{\mu}-Y_M)+Y_M\right ) \\
=& \frac{\widetilde{\sigma}}{s_1} \int_{\widetilde{\mu}=Y_{M}-\epsilon }^{Y_{M}}\left(\frac{Y_M-\widetilde{\mu}}{s_1}+exp\left(-\frac{Y_M-\widetilde{\mu}}{s_1}\right)-1\right)\ \mathrm{d}\widetilde{\mu} \\
=&\left.\frac{\widetilde{\sigma}}{s_1}\left(-\frac{s_1}{2} \left (\frac{Y_M-\widetilde{\mu} }{s_1} \right )^2+s_1\cdot exp\left( -\frac{Y_M-\widetilde{\mu}}{s_1} \right )-\widetilde{\mu}\right)\right|^{Y_M}_{Y_{M}-\epsilon}\\
=& \widetilde{\sigma}\left(\frac{\epsilon ^2}{2s_1^2} -\frac{\epsilon}{s_1}-exp\left (-\frac{\epsilon }{s_1} \right) +1\right)
\end{aligned}
\end{equation}

The area integral of $R_2$ is as follows:

\begin{equation}
\label{equ:a14}
\begin{aligned}
&\int_{\widetilde{\mu}_s=Y_{M}}^{Y_{M}+\frac{\widetilde{\sigma}}{s_2}\epsilon}obj_M\left(\widetilde{\mu}_s\right ) \ \mathrm{d}\widetilde{\mu}_s,\ \widetilde{\mu}_s=\frac{\widetilde{\sigma}}{s_2}  (\widetilde{\mu}-Y_M)+Y_M \\
=&\int_{\widetilde{\mu}=Y_{M}}^{Y_{M}+\epsilon} \left( \frac{Y_{M}-\left (\frac{\widetilde{\sigma}}{s_2} (\widetilde{\mu}-Y_M)+Y_M\right )}{\widetilde{\sigma}}+exp\left(-\frac{Y_{M}-\left (\frac{\widetilde{\sigma}}{s_2} (\widetilde{\mu}-Y_M)+Y_M\right )}{\widetilde{\sigma}}\right)-1\right)\ \mathrm{d}\left (\frac{\widetilde{\sigma}}{s_2} (\widetilde{\mu}-Y_M)+Y_M\right ) \\
=& \frac{\widetilde{\sigma}}{s_2} \int_{\widetilde{\mu}=Y_{M} }^{Y_{M}+\epsilon}\left(\frac{Y_M-\widetilde{\mu}}{s_2}+exp\left(-\frac{Y_M-\widetilde{\mu}}{s_2}\right)-1\right)\ \mathrm{d}\widetilde{\mu} \\
=&\left.\frac{\widetilde{\sigma}}{s_2}\left(-\frac{s_2}{2} \left (\frac{Y_M-\widetilde{\mu} }{s_2} \right )^2+s_2\cdot exp\left( -\frac{Y_M-\widetilde{\mu}}{s_2} \right )-\widetilde{\mu}\right)\right|^{Y_M+\epsilon}_{Y_{M}}\\
=& \widetilde{\sigma}\left(-\frac{\epsilon ^2}{2s_2^2} -\frac{\epsilon}{s_2}+exp\left (\frac{\epsilon }{s_2} \right) -1\right)
\end{aligned}
\end{equation}

In order to achieve equality between the areas of $R_1$ and $R_2$, we obtain the following equation:

\begin{equation}
\label{equ:a15}
\begin{aligned}
&\widetilde{\sigma}\left(\frac{\epsilon ^2}{2s_1^2} -\frac{\epsilon}{s_1}-exp\left (-\frac{\epsilon }{s_1} \right) +1\right)=\widetilde{\sigma}\left(-\frac{\epsilon ^2}{2s_2^2} -\frac{\epsilon}{s_2}+exp\left (\frac{\epsilon }{s_2} \right) -1\right)\\
&\frac{\epsilon ^2}{2s_1^2}+\frac{\epsilon ^2}{2s_2^2} -\frac{\epsilon}{s_1}+\frac{\epsilon}{s_2}-exp\left (-\frac{\epsilon }{s_1} \right)-exp\left (\frac{\epsilon }{s_2} \right) +2 = 0
\end{aligned}
\end{equation}

The parameter $\epsilon$ is associated with the prediction accuracy. The higher the prediction accuracy is, the smaller the error is. The closer the predicted $\widetilde{\mu}$ will be to the real target $Y_M$, the smaller the area we need to consider, that is, the smaller $\epsilon$ will be. On the contrary, when the prediction accuracy is very low, the $\widetilde{\mu}$ predicted by the model is relatively far away from $Y_M$, so the $\epsilon$ we need to consider must be appropriately increased. 

In the actual training, we select $\epsilon=0.1\widetilde{\sigma}$ as our choice, because of the observation that the relative error of the model's predictions for extreme values is often below 10\% ~\cite{bellprat2016attribution} ~\cite{fang2021survey}. Therefore it is sufficient to consider an error margin of 10$\%$. On the other hand, if the error range considered is too small, the scaling effect will not be obvious. When we set $\epsilon=0.1\widetilde{\sigma}$, the solution to Equation \ref{equ:a15} remains non-unique. Since our primary interest lies in the relative relationship between $s_1$ and $s_2$, it is reasonable to fix one of them first. We fix $s_2$ to $\widetilde{\sigma}$, and calculate that $s_1=0.9\widetilde{\sigma}$ approximately.

Thus we can obtain the specific form of the scaling function, depicted below:

\begin{equation}
\label{equ:a16}
\widetilde{\mu}_s=\left\{\begin{matrix}

\begin{aligned}
&\frac{10}{9} (\widetilde{\mu}-Y_M)+Y_M,\ \widetilde{\mu}\le Y_M \\
&\widetilde{\mu},\ \widetilde{\mu}>  Y_M  \\

\end{aligned}
\end{matrix}\right.
\end{equation}

Subtract $Y_M$ from both sides to get:

\begin{equation}
\label{equ:a17}
\begin{aligned}
\widetilde{\mu}_s-Y_M&=\left\{\begin{matrix}
\frac{10}{9} (\widetilde{\mu}-Y_M),\ \widetilde{\mu}\le Y_M \\
\widetilde{\mu}-Y_M,\ \widetilde{\mu}>  Y_M  \\
\end{matrix}\right.\\
(\widetilde{\mu}_s-Y_M)^2&=\left\{\begin{matrix}
\frac{100}{81} (\widetilde{\mu}-Y_M)^2,\ \widetilde{\mu}\le Y_M \\
(\widetilde{\mu}-Y_M)^2,\ \widetilde{\mu}>  Y_M  \\
\end{matrix}\right.
\end{aligned}
\end{equation}

Similarly, for the prediction of minimum values $Y_m$, we can also apply a similar scaling function:

\begin{equation}
\label{equ:a18}
(\widetilde{\mu}_s-Y_m)^2=\left\{\begin{matrix}

\begin{aligned}
&\frac{100}{81} (\widetilde{\mu}-Y_m)^2,\ \widetilde{\mu}\ge  Y_m \\
&(\widetilde{\mu}-Y_m)^2,\ \widetilde{\mu}<  Y_m  \\

\end{aligned}
\end{matrix}\right.
\end{equation}

Finally, merge the two scaling functions into $S(\hat{Y},Y)$:

\begin{equation}
\label{equ:a19}
S(\hat{Y},Y)=\left\{\begin{matrix}

\begin{aligned}
&\frac{100}{81} ,\ (\hat{Y} \ge Y \ and\ Y < Y_{10th}) \ or\  (\hat{Y} \le Y \ and\ Y > Y_{90th}) \\
&1,\ else  \\

\end{aligned}
\end{matrix}\right.
\end{equation}

where $\hat{Y}$ is the prediction and $Y$ is the target. $Y_{10th}$ and $Y_{90th}$ correspond to the 10th and 90th percentiles in the sample, respectively, and they can serve as thresholds for extreme values (such as extremely low or high temperatures). The expression $(\hat{Y} \ge Y \ and\ Y < Y_{10th}) \ or\  (\hat{Y} \le Y \ and\ Y > Y_{90th})$ refers to the situation where the true value meets the threshold of extreme values, but the model gives an underestimated prediction. By applying the scaling function $S(\cdot)$ to the loss function, we obtain Exloss. It is noticeable that the model increases the loss when underestimating extremes, thereby imposing a higher penalty.

\begin{equation}
\label{equ:a20}
Exloss\left (\hat{\mathcal Y},\mathcal Y\right )=\left \| S(\hat{\mathcal Y},\mathcal Y)\odot \left(\hat{\mathcal Y}- \mathcal Y\right) \right \|^2
\end{equation}

\section{ExBooster}
\label{C}

The python code of ExBooster (PyTorch implementation) is implemented as follows:

\begin{python}
def SortIndex(input):
    '''
    Get the sorting ranking corresponding to each element in tensor.
    Args:
        input (torch.Tensor): The input tensor.
    Returns:
        torch.Tensor: Tensor of the same size as input. The ranking of each element in the tensor, from smallest to largest.
    Example:
        input=torch.tensor([1.2, 1.5, 0.8, 0.9])
        return torch.tensor([2, 3, 0, 1])
    '''
    sorted_indices = torch.argsort(input, dim=-1)
    ranks = torch.argsort(sorted_indices, dim=-1)
    return ranks
\end{python}

\begin{python}
def ExBooster(pred, sampling_nums=50, noise_scale=0.1):
    '''
    Apply ExBooster to the predictions.
    Args:
        pred (torch.Tensor): Tensor of size [B, C, H, W]. The input predictions.
        sampling_nums (int): Number of samplings (default: 50).
        noise_scale (float or torch.Tensor): Scaling factor for the noise. It can be a real number or a tensor of size [B, C, H, W] (default: 0.1).
    Returns:
        torch.Tensor: Tensor of size [B, C, H, W]. The extreme boosted predictions.
    '''
    B, C, H, W = pred.shape

    scale = noise_scale * torch.ones_like(pred) # [B, C, H, W]

    # Get index
    idx = SortIndex(pred.flatten(2,3)) # [B, C, H*W]
    
    # Sample
    pred = pred.unsqueeze(2) # [B, C, 1, H, W]
    scale = scale.unsqueeze(2) # [B, C, 1, H, W]
    disturbance = torch.randn(B, C, sampling_nums, H, W, device=pred.device) * scale 
    ens = pred + disturbance # [B, C, sampling_nums, H, W]

    # Sort ensembles
    sorted_ens, _ = torch.sort(ens.flatten(2,4)) # [B, C, sampling_nums*H*W]
    sorted_ens = sorted_ens.reshape(B, C, H*W, sampling_nums) # [B, C, H*W, sampling_nums]

    # Partition and median
    k = int(0.5 * sampling_nums) # sampling_nums / 2
    sorted_ens_mid, _ = torch.kthvalue(sorted_ens, k, -1) # [B, C, H*W]

    # Restore by index
    ens_from_idx = torch.gather(sorted_ens_mid, dim=-1, index=idx) # [B, C, H*W]
    out = ens_from_idx.reshape(B, C, H, W) # [B, C, H, W]
    
    return out
\end{python}

\section{Training Details}
\label{D}
Our model is trained for 2 weeks using 32 A100 GPUs, and the batch size is 32. The number of learnable parameters of $M_d$ are 570M, the learning rate of stage 1 is 2.5e-4, and the learning rate of stage 2 is 5e-6. The number of learnable parameters of $M_g$ are 140M, and the learning rate is 5e-4. The complete hyperparameter settings are shown in the table below.

\begin{table*}[h]
\centering
\begin{tabular}{ll}
\toprule
 Hyperparameter & Value  \\
\midrule
Batch size & 32 \\
Learning rate of stage 1 & 2.5e-4 \\
Learning rate of stage 2 & 5e-6 \\
Learning rate of stage 3 & 5e-4 \\
Learning rate schedule & Cosine \\
Optimizer & AdamW \\
Patch size & 8x8 \\
Embedding dimension	& 1152 \\
Attention type & Swin window \\
Attention heads & 6 \\
Window size & 6x12 \\
Dropout & 0 \\
MLP ratio & 4 \\
Activation function & GLUE \\
Data size & [B,69,721,1440] \\
\bottomrule
\end{tabular}
\caption{\textbf{Hyperparameter.}}
\end{table*}

The process (including training and testing) of our model can be divided into four distinct stages:

\begin{itemize}
\label{sub:stages}
\item Stage 1: One-Step pretraining. Use MSE loss to train $M_d$ learning mapping from $\mathcal X^{i}$ to $\mathcal X^{i+1}$ as existing data-driven global weather forecasting models.
\item Stage 2: Muti-Step finetuning. Use the proposed Exloss to finetune $M_d$ learning autoregressive forecasting. This stage improves not only the accuracy of the model's multi-step predictions, but also the ability to predict extreme weather due to the use of Exloss.
\item Stage 3: Diffusion training. Use Exloss to train $M_g$ learning adding details to the output of $M_d$ while freezing the parameters of $M_d$.
\item Stage 4: Inference. In this stage, the ExBooster module that does not require training will be added for complete weather forecast.
\end{itemize}

We use Exloss in Stage 2 and Stage 3 of model training.

In Stage 2, the model will perform multi-step autoregressive fineutne on $M_d$ through the replay buffer ~\cite{chen2023fengwu}. We employ Exloss as the loss function in this stage to help the model achieve accurate multi-step predictions for extreme weather events.

In Stage 3, We freeze the parameters of $M_d$ and train the diffusion model $M_g$ to generate the land-surface variables (u10, v10, t2m) with more details. In order to further improve the expression of diffusion for extreme values, we apply Exloss in diffusion to directly optimize extreme values and avoid losing details as prediction time increases.

% Specifically, first train a deterministic model through Exloss, denoted as $M_d(\mathcal X^i)=\mathcal X^{i+1}_{d}$. Subsequently, train a diffusion model using $\mathcal X^{i+1}_{d}$ as the condition to generate the land-surface variables (u10, v10, t2m) with more details, denoted as $M_g(\mathcal X^{i+1}_{d})=\mathcal X^{i+1}_{g}$. Finally, we concatenate not-land-surface variables (msl, z, q, u, v, t) of $\mathcal X^{i+1}_{d}$ with land-surface variables of $\mathcal X^{i+1}_{g}$ to obtain more detailed forecasts $\mathcal X^{i+1}$.

% In order to specifically strengthen the expression of extreme values, we also added the Exloss term to the diffusion training loss.

\begin{equation}
\label{equ:11}
Loss(\hat{\varepsilon}, \varepsilon )=\left \| \hat{\varepsilon}- \varepsilon \right \| ^2+Exloss(\hat{\varepsilon}, \varepsilon)
\end{equation}

where $\hat{\varepsilon}$ and $\varepsilon$ represent the predicted Gaussian noise ~\cite{yang2023diffusion} and the actual added noise in diffusion. 

Finally, concatenate the not-land-surface variables (msl, z, q, u, v, t) of $M_d$'s output with the land-surface variables of $M_g$'s output to obtain more detailed forecasts.

\section{Module Processing Visualization}
\label{E}
As described in Section \ref{sub:stages}, our model comprises two trainable modules, namely the deterministic model (denoted as $M_d$) and the probabilistic generation model (denoted as $M_g$), along with the training-free ExBooster module. The outputs of these modules are denoted as $P_d$, $P_g$, and $P_e$ respectively. By visualizing the difference between $P_g$ and $P_d$, we can observe the impact of the diffusion model, while visualizing the difference between $P_e$ and $P_g$ allows us to examine the effect of the ExBooster module.

\begin{figure}[ht]
    \centering
    \begin{minipage}[b]{0.68\textwidth}
        \centering
        \includegraphics[width=\textwidth]{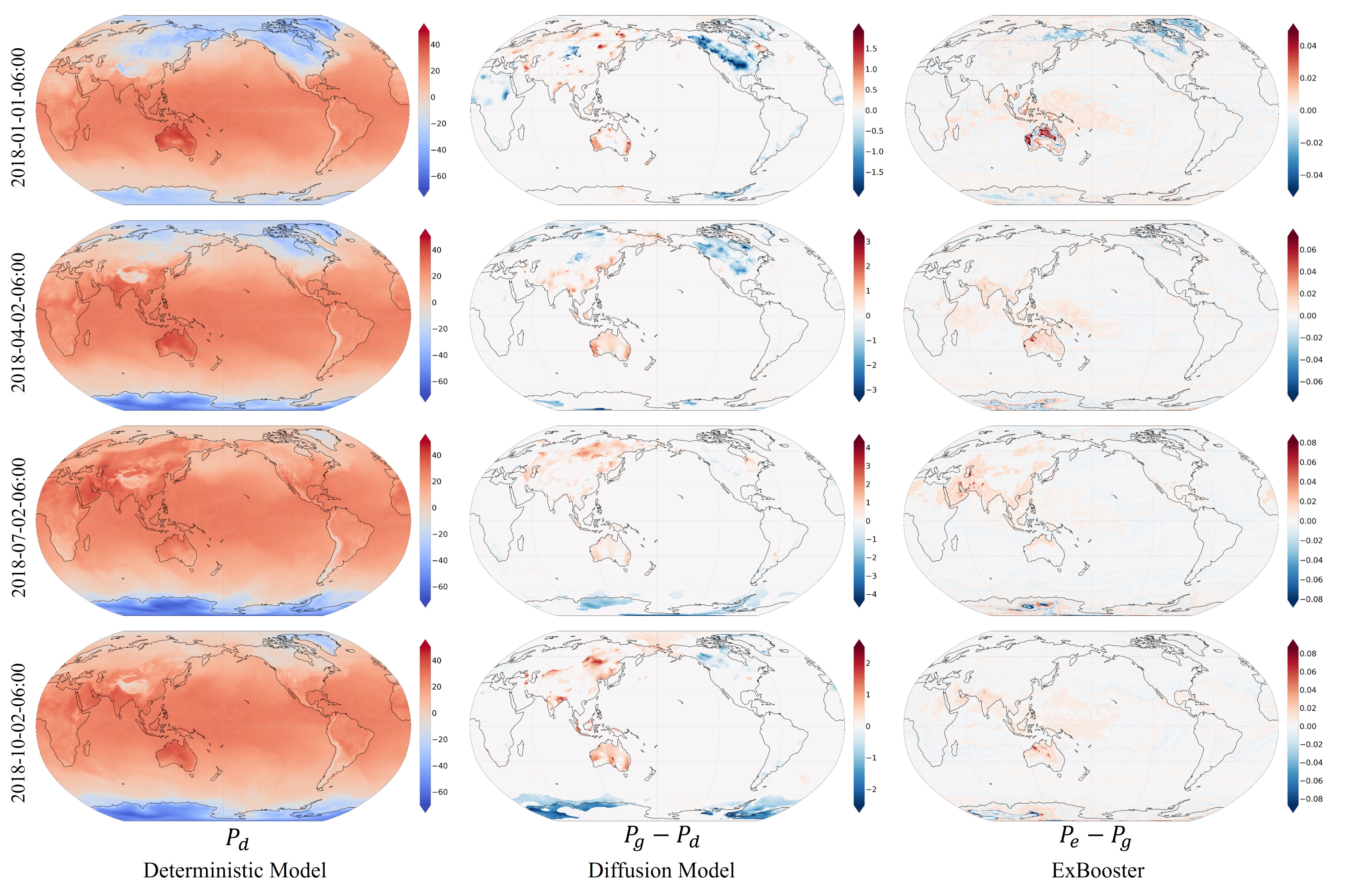}
        \vspace{-0.2cm}
        \caption{\textbf{Processing of Different Modules of t2m.}}
        \label{fig: t2m_process}
    \end{minipage}
    \vfill
    \begin{minipage}[b]{0.68\textwidth}
        \centering
        \includegraphics[width=\textwidth]{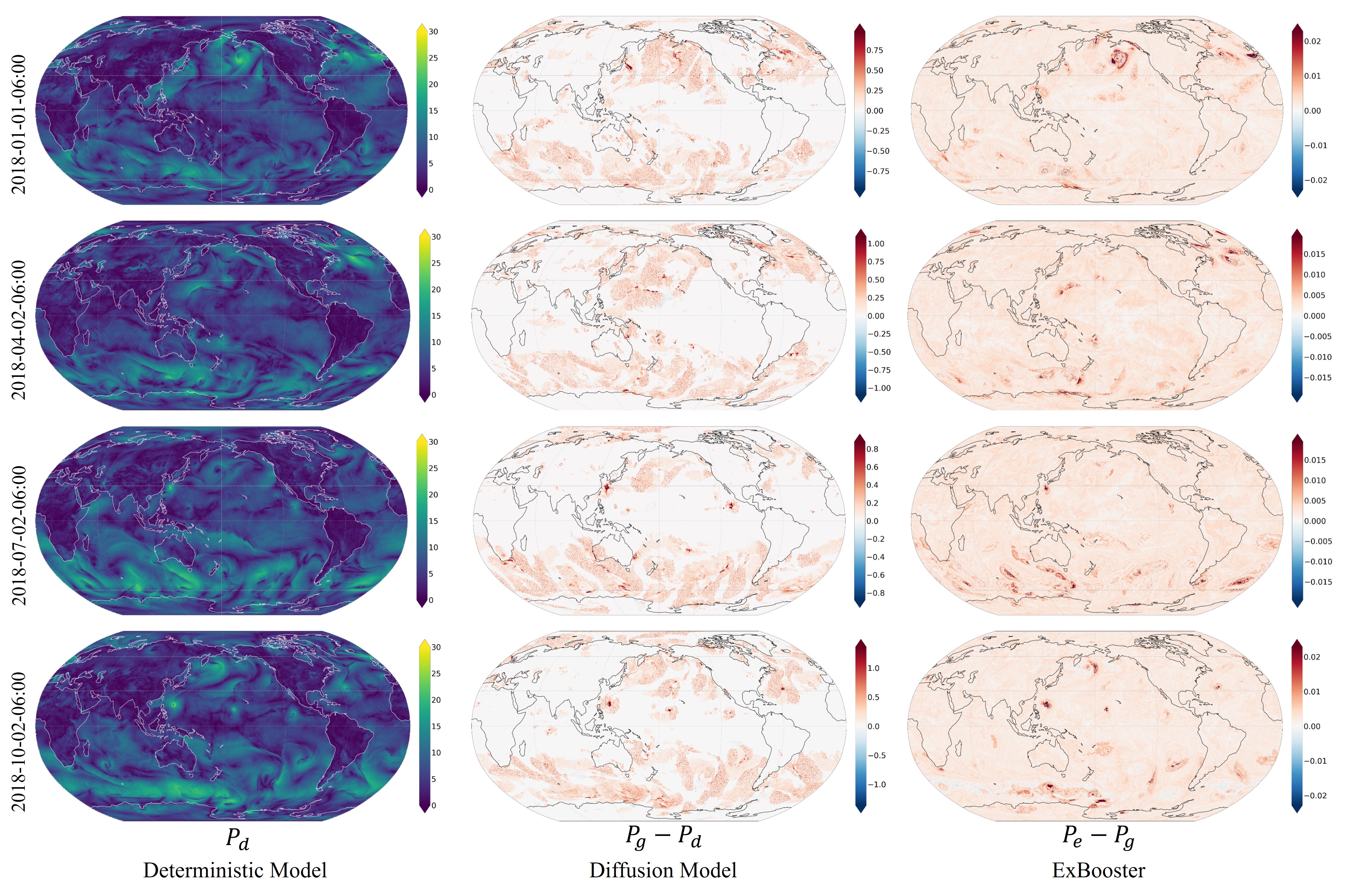}
        \vspace{-0.2cm}
        \caption{\textbf{Processing of Different Modules of ws10.}}
        \label{fig: ws10_process}
    \end{minipage}
    \label{fig:main}
\end{figure}

Two key observations can be derived from Figure \ref{fig: t2m_process}. Firstly, both the diffusion and ExBooster modules exert similar influences on the final output. This indicates that our distinct modules show consistency and do not adversely affect one another. Secondly, both the diffusion and ExBooster modules result in increased temperatures in mid-latitudes and decreased temperatures in high-latitudes ~\cite{richard2013temperature} ~\cite{francis2012evidence}. Collectively, this contributes to a certain level of extreme weather events at the global scale.

As depicted in Figure \ref{fig: ws10_process}, both the diffusion and ExBooster modules exhibit an amplifying effect on wind speed, consequently leading to an increase in extreme wind speed. Furthermore, in the vicinity of cyclones ~\cite{gray1977tropical} exhibiting higher wind speeds, we observe a pronounced intensification of the enhancement effect facilitated by the diffusion and ExBooster modules. This notable enhancement is particularly advantageous in the accurate prediction of typhoons and their associated wind speeds ~\cite{wei2018regional} ~\cite{yang2019using}.

\end{document}